%% file: main.tex
\documentclass[
  journal=medium,
  manuscript=technical-report,
  year=2026,
  volume=0,
]{cup-journal}
\usepackage{amsmath}
\usepackage[nopatch]{microtype}
\usepackage{booktabs}
\usepackage{xspace}
\usepackage{amssymb}

\usepackage[T1]{fontenc}    %
\usepackage{hyperref}       %
\usepackage{url}            %
\usepackage{amsfonts}       %
\usepackage{eso-pic}        %
\usepackage{nicefrac}       %
\usepackage{fancyhdr}       %
\usepackage{graphicx}
\usepackage{multirow}

\colorlet{tbrowcolor}{white}
\colorlet{tbheadcolor}{white}

\title{HASTE: A Platform for Rapid \\ Post-Disaster Building Damage Assessment}

\author{Caleb Robinson\textsuperscript{\textdagger}}
\affiliation{Microsoft AI for Good Research Lab}
\email[Author1]{caleb.robinson@microsoft.com}

\author{Anthony Ortiz\textsuperscript{\textdagger}}
\affiliation{Microsoft AI for Good Research Lab}
\email[Author2]{anthony.ortiz@microsoft.com}

\author{Simone Fobi Nsutezo}
\affiliation{Microsoft AI for Good Research Lab}

\author{Cameron Birge}
\affiliation{Microsoft AI for Good Research Lab}

\author{Meygha Machado}
\affiliation{Microsoft AI for Good Research Lab}

\author{Marcelo Duarte}
\affiliation{Microsoft AI for Good Research Lab}

\author{Joaquin Rivero Rodriguez}
\affiliation{Microsoft AI for Good Research Lab}

\author{Anthony Cintron Roman}
\affiliation{Microsoft AI for Good Research Lab}

\author{Kevin White}
\affiliation{Microsoft AI for Good Research Lab}

\author{Inbal Becker-Reshef}
\affiliation{Microsoft AI for Good Research Lab}

\author{Juan M. Lavista Ferres}
\affiliation{Microsoft AI for Good Research Lab}

\makeatletter
\patchcmd{\@maketitle}
  {\parbox[b]{\dimexpr\textwidth-26mm\relax}}
  {\parbox[b]{\dimexpr\textwidth-56mm\relax}}
  {}{}
\patchcmd{\@maketitle}
  {\includegraphics[width=26mm]{figures/ai4g-logos}}
  {\raisebox{-2.5mm}{\includegraphics[width=54mm]{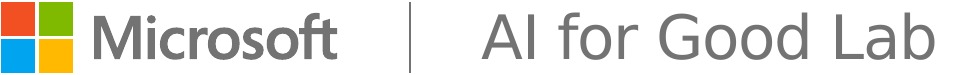}}}
  {}{}
\patchcmd{\@maketitle}
  {{\volumefont\cup@vol}, }
  {}
  {}{}
\renewcommand\paragraph{\@startsection{paragraph}{4}{\z@}%
  {3.25ex \@plus1ex \@minus.2ex}%
  {-1em}%
  {\sffamily\bfseries}}
\pretocmd{\cup@contact@details}%
  {{\textsuperscript{\textdagger}These authors contributed equally.}\par}%
  {}{}
\makeatother

\begin{document}

\maketitle

\begin{abstract}
When a large disaster strikes, responders need a map of which buildings are damaged within hours.
The models that do well on public benchmarks assume matched before-and-after imagery and a training set drawn from similar past events, and neither is usually available for a new disaster in its first day.
We present HASTE (High-speed Assessment and Satellite Tracking for Emergencies), a no-code web platform that lets analysts who are not machine learning engineers produce per-building damage maps from post-disaster satellite imagery.
HASTE implements two methods that share one interface.
The first requires the user to label polygons over the post-disaster scene, trains a small semantic segmentation model on that single scene, runs it over the whole image, and joins the per-pixel output to existing building footprints.
The second embeds every footprint with a pretrained vision model, requires the user to label a handful of buildings, and fits a logistic regression in the browser that scores the rest of the scene in seconds.
We describe the platform, both methods, and the engineering that supports them.
We also report preliminary experiments on xBD showing that foundation-model embeddings pooled over footprints separate damaged from intact buildings using post-disaster imagery alone, matching a fully supervised ResNet-50 baseline with a twentieth of its labels.
HASTE and its predecessors have supported more than thirty real-world disaster responses since 2023, spanning earthquakes, hurricanes, cyclones, floods, wildfires, and tornadoes, delivering results to humanitarian partners within hours to days of imagery becoming available.
We close with the directions we think are most promising, including vision-language assessment, active learning, and damage models for roads and other infrastructure.
HASTE is open source at \url{https://github.com/microsoft/haste}.
\end{abstract}

\section{Introduction}

Earthquakes, hurricanes, floods, and wildfires damage or destroy buildings faster than any ground survey can count them.
In the first hours after an event, humanitarian organizations decide where to send search-and-rescue teams, how much shelter to stage, and how to route aid, and these decisions depend on knowing which neighborhoods took the worst damage \citep{ceferino2024placing}.
Satellite and aerial imagery can cover a whole city in a single pass, so it has become the main source of this early picture.
Turning that imagery into a building-by-building damage map, quickly and reliably, is the problem HASTE addresses.

\begin{figure}[t!h]
\centering
\includegraphics[width=0.9\textwidth]{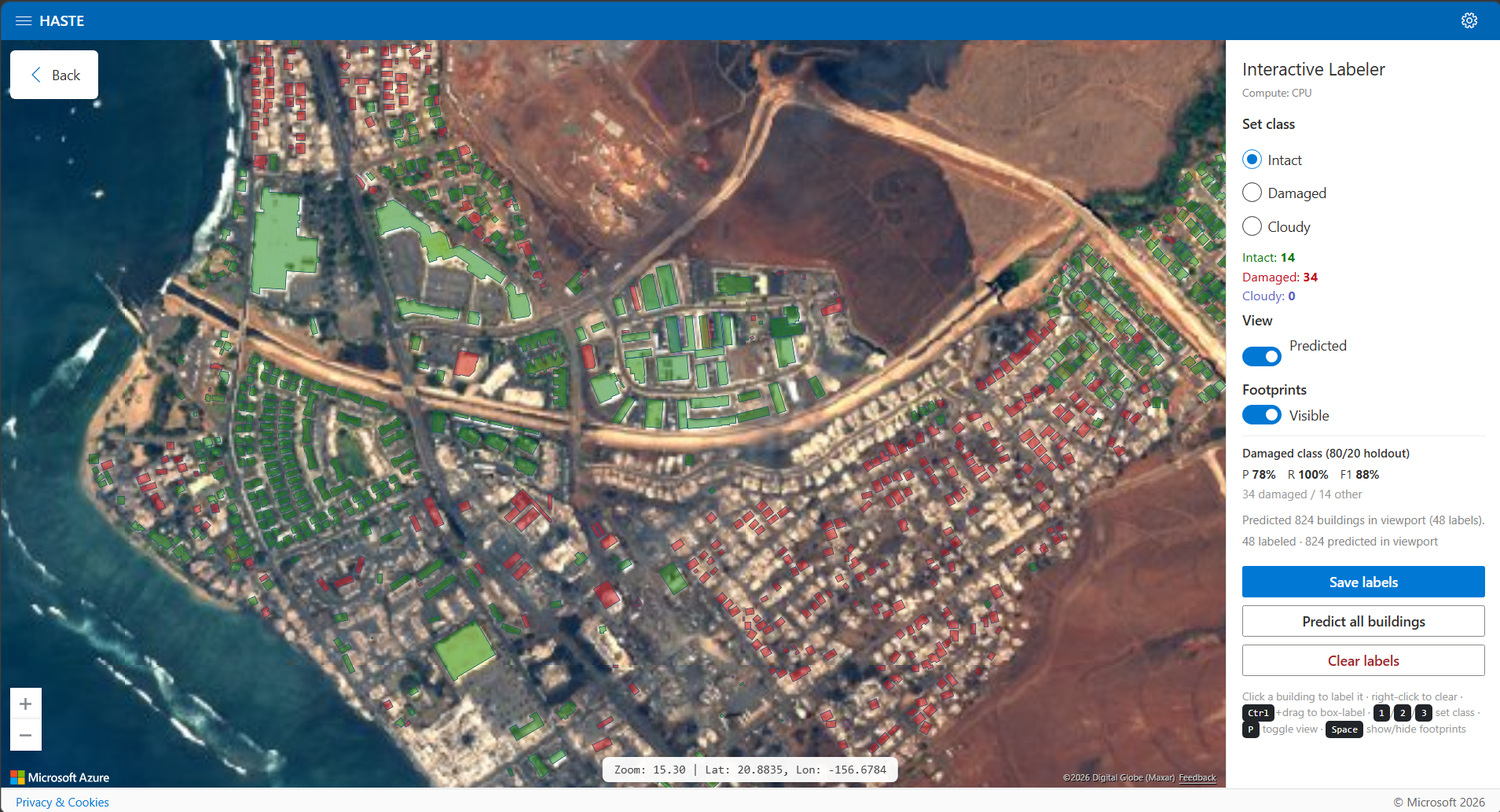}\\[1pt]
\includegraphics[width=0.9\textwidth]{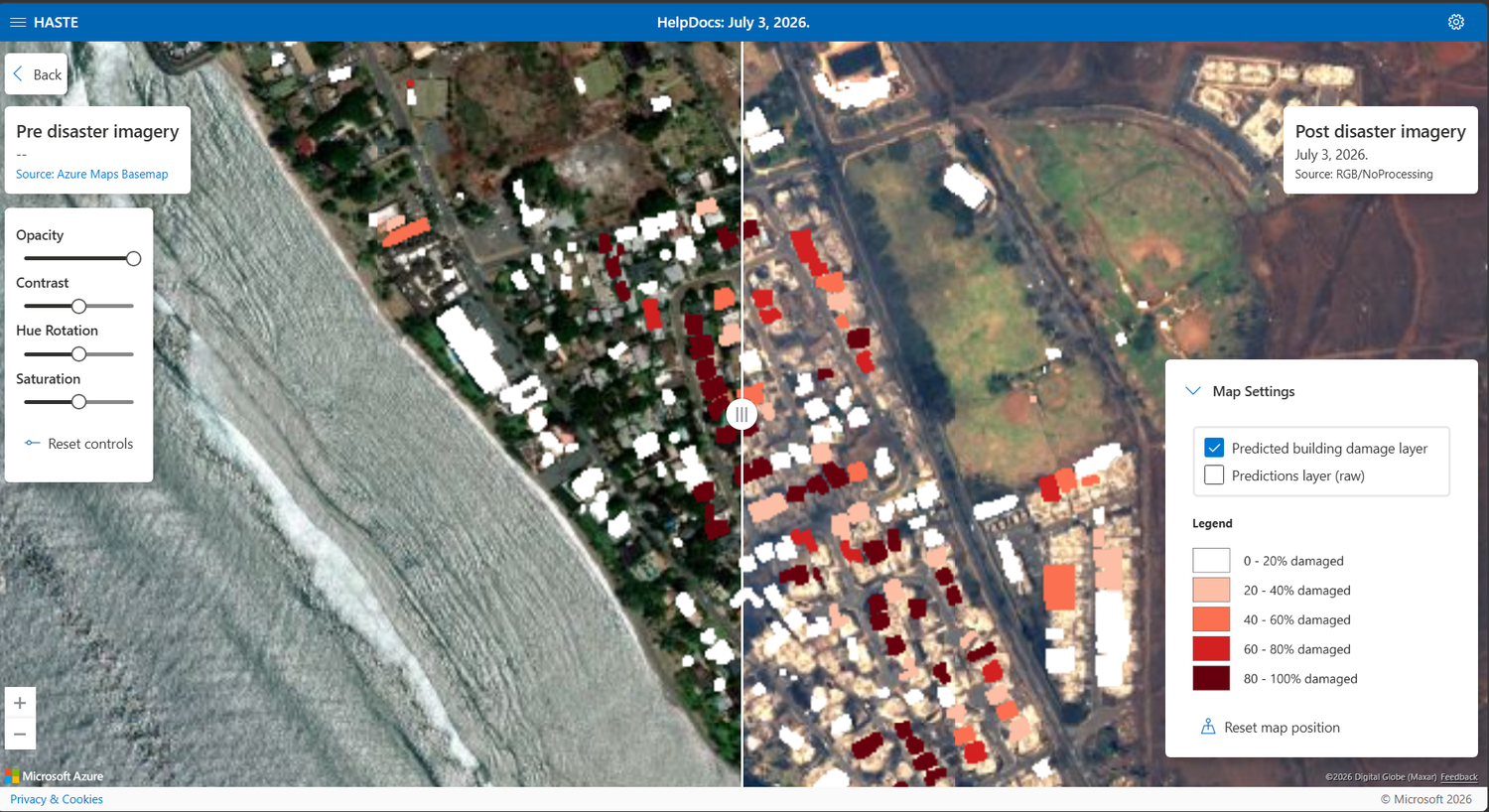}
\caption{HASTE in use.
Top, the interactive labeler over the 2023 Lahaina wildfire scene: the analyst has labeled 48 buildings, and the in-browser model scores every footprint in view (green intact, red damaged) alongside holdout metrics for the damaged class.
Bottom, the results visualizer, with pre- and post-disaster imagery in a swipe view and the predicted per-building damage layer binned by damage fraction.}
\label{fig:app}
\end{figure}

Machine learning has made steady progress on damage assessment, driven largely by the xBD dataset and the xView2 challenge \citep{gupta2019xbd,xview2}.
The strongest models read a pre-event and a post-event image of the same place and predict a damage level for every pixel or building \citep{weber2020building,adriano2021learning,kaur2023large}.
This works well when the test event resembles the training events and when clean imagery from both dates is available.
A real response rarely offers either.
Pre-event imagery may be years old or missing, the new disaster may look nothing like the ones in the training set, and the sensor, resolution, and viewing angle change from one response to the next.
A study of transfer across 13 disasters confirms that a model trained on past events loses accuracy on a new one \citep{valentijn2020multi}.
At the same time, labels for the new event are scarce in the window when they would matter most.

Operational services work around these gaps with human effort.
The Copernicus Emergency Management Service and the UNITAR/UNOSAT rapid mapping service produce maps by expert interpretation.
The Humanitarian OpenStreetMap Team coordinates volunteers to trace features by hand \citep{copernicusems,lang2017earth,hotosm}.
Both approaches are valuable, and both are hard to scale to every building in a large city on the first day.
The gap that remains is a way to get an accurate, building-level map fast, with only a small amount of local labeling, run by the analysts who are already doing the response rather than by a separate machine learning team.

HASTE fills that gap by training a small model for each event from a handful of labels the analyst draws on the post-disaster scene, and it wraps the whole process in a web interface with no code (Figure~\ref{fig:app}).
It grew out of several years of work at the Microsoft AI for Good Research Lab \citep{gholami2022deployment,robinson2022fast,robinson2023rapid,robinson2023turkey} and turns those one-off response pipelines into a platform that a partner organization can run on its own.
This report describes the platform, the two methods it offers, the responses it has supported, and where we plan to take it next.

Our contributions are the following.
We describe HASTE, a deployed no-code platform for rapid building damage assessment, and its architecture.
We lay out its two assessment methods and the trade-off between them, one based on per-scene segmentation and one on footprint embeddings with an in-browser classifier.
We report preliminary experiments on xBD that benchmark how per-building damage accuracy scales with the amount of labeled data and how it compares against a supervised baseline.
We focus these results on the footprint-embedding method because its in-browser classifier retrains in seconds, letting an analyst see the effect of each label immediately and iterate far faster than the segmentation method allows.
Finally, we summarize the lab's operational response history and the open problems we consider most worth pursuing.

\section{Related work}

\paragraph{Datasets and benchmarks.}
The xBD dataset and the xView2 challenge set the terms for this field \citep{gupta2019xbd,gupta2019creating,xview2}.
xBD pairs pre- and post-event very-high-resolution imagery over nineteen disasters with more than 850,000 labeled building polygons on a four-level damage scale \citep{gupta2019xbd}.
SpaceNet established the earlier building-footprint extraction task \citep{van2018spacenet}, and BRIGHT adds radar so that responders can map damage through cloud cover \citep{chen2025bright}.

\paragraph{Damage-assessment models.}
Most competitive xView2 systems use a two-branch network that compares the pre- and post-event images.
Weber and Kané fuse the two dates and report competitive results on the xView2 leaderboard \citep{weber2020building}, DAHiTrA applies a hierarchical transformer \citep{kaur2023large}, Adriano et al.\ combine optical and radar across time \citep{adriano2021learning}, and RescueNet couples building segmentation with damage classification in a single network \citep{gupta2021rescuenet}.
These models are accurate on xBD, and they assume the matched pre/post imagery and in-distribution training data that a live response often lacks.
That gap is what pushed our own work toward per-event training, and it has pushed others toward single-temporal methods that relax the need for matched pre/post imagery \citep{zheng2024towards,wang2025cross}.
Gholami et al.\ built an early deployable tool with a multitask Siamese U-Net that traded accuracy for a large gain in speed \citep{gholami2022deployment}, and later work showed that a model overfit to one scene needs only a few hundred local labels to extract footprints well \citep{robinson2022fast}, the label-efficiency idea behind HASTE's first method.
The Turkey earthquake report applied this per-event approach at scale after the February 2023 earthquakes \citep{robinson2023turkey}, a doublet whose building damage has since been assessed in detail \citep{chen2025assessing}.
The Rolling Fork tornado response the following month put the human-in-the-loop per-event workflow into practice, reaching 0.86 precision and 0.80 recall against field ground truth in under two hours per scene \citep{robinson2023rapid}.
An independent study later benchmarked the Turkey model for health-facility damage and found high specificity with low sensitivity \citep{ramachandran2025implementation}, which we return to in Section~\ref{sec:responses}.

\paragraph{Foundation models and embeddings.}
HASTE's second method rests on pooling features from a pretrained vision model over each building and fitting a small classifier.
MOSAIKS is the simplest version, using random convolutional features with a linear model \citep{rolf2021generalizable}.
Stronger features come from self-supervised backbones such as DINOv2 and DINOv3 \citep{oquab2023dinov2,simeoni2025dinov3}, a DINOv2 variant pretrained on satellite imagery for canopy-height mapping \citep{tolan2024very}, and geospatial foundation models such as Prithvi and SatMAE \citep{jakubik2023foundation,cong2022satmae}.
A recent review covers the broader decade of building damage assessment with machine learning and remote sensing \citep{al2024integrating}.

\paragraph{Interactive and embedding-based assessment.}
Two ideas underpin HASTE's second method, and each has precedent on its own.
The first is fitting a small model on frozen features.
MOSAIKS computes task-agnostic satellite features once and trains a linear model per task \citep{rolf2021generalizable}, and recent work brings pretrained vision transformers to building damage, comparing frozen and end-to-end fine-tuned variants \citep{siva2026building} or pooling backbone features over footprints with a task-specific head \citep{xiao2026damage}.
These run offline as batch jobs.
The second idea is interactive labeling with immediate model feedback.
Our earlier land-cover work retrained on frozen deep features fast enough to feel live in the browser \citep{robinson2020human}, and foundation-model segmentation tools have since made interactive object delineation routine \citep{kirillov2023segment}.
Combining the two for post-disaster damage, pooling frozen embeddings over individual footprints and training a classifier that an analyst retrains and re-scores in seconds, is uncommon next to the per-pixel segmentation that dominates work on xBD.
HASTE's second method sits in that gap, and the experiments in Section~\ref{sec:experiments} probe its design.

\section{The HASTE platform}
\label{sec:tool}

HASTE is a web platform that takes a partner from raw post-disaster imagery to a per-building damage map and a summary report, without writing code.
A project holds one or more image layers.
For each layer the analyst chooses one of the two methods described in the next sections, does a small amount of labeling, and reviews the result on an interactive map.

\begin{figure}[t]
\centering
\includegraphics[width=0.86\textwidth]{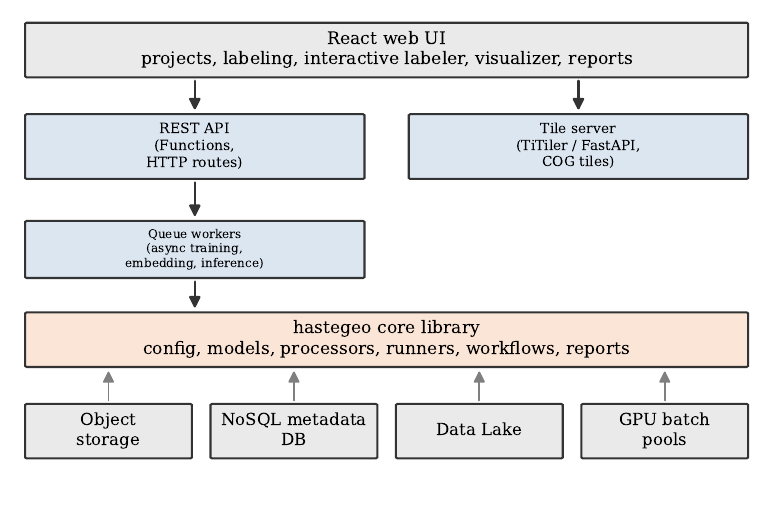}
\caption{HASTE system architecture.
A React web interface talks to a REST API and a tile server.
The API places long-running work on queues, and workers pick up the jobs and call the shared \texttt{hastegeo} core library.
The library runs training, embedding, and inference jobs on GPU compute and reads and writes imagery, footprints, models, and results from cloud storage.}
\label{fig:architecture}
\end{figure}

\paragraph{Architecture.}
Figure~\ref{fig:architecture} shows the main pieces.
A React single-page application provides project management, the two labeling tools, the map visualizer, and the reports.
It talks to a REST API and to a tile server that streams cloud-optimized GeoTIFFs to the map.
When an analyst starts a training, embedding, or inference job, the API drops a message on a queue, and a separate pool of workers picks it up.
The workers call a shared core library, \texttt{hastegeo}, that holds the configuration, data models, processors, job runners, and the scripted workflows that do the heavy work.
Jobs that need a GPU run on an elastic batch pool.
Imagery, building footprints, trained models, embeddings, and prediction layers live in cloud storage, and project and job metadata live in a database.
This split keeps the interface responsive while model training and full-scene inference run in the background.

\paragraph{Imagery and footprints.}
Before either method runs, HASTE prepares the scene.
It mosaics the post-disaster imagery into a cloud-optimized GeoTIFF and pulls existing building footprints for the area from open sources such as the Microsoft and Overture footprint datasets \citep{microsoft2022footprints,overture2024}.
Both methods produce their final answer as a value attached to each of these footprints, so the footprint layer is the common reference layer for the scene.

\paragraph{Reports.}
Whichever method produced them (Sections~\ref{sec:method1} and~\ref{sec:method2}), the per-building predictions feed a shared validation and assessment report.
An analyst can draw a validation sample and label those buildings by visual interpretation of the imagery, and the report then computes accuracy, precision, recall, and average precision against that sample, along with a precision--recall curve and a confusion matrix.
Because a sample rarely covers every building, the report also gives a population estimate of the total number of damaged buildings in the scene with a 95\% confidence interval, computed with a finite-population correction.
This lets a partner attach an uncertainty to the headline count they report upward.

\section{Method 1: segmentation and footprint join}
\label{sec:method1}

\begin{figure}[t]
\centering
\includegraphics[width=0.92\textwidth]{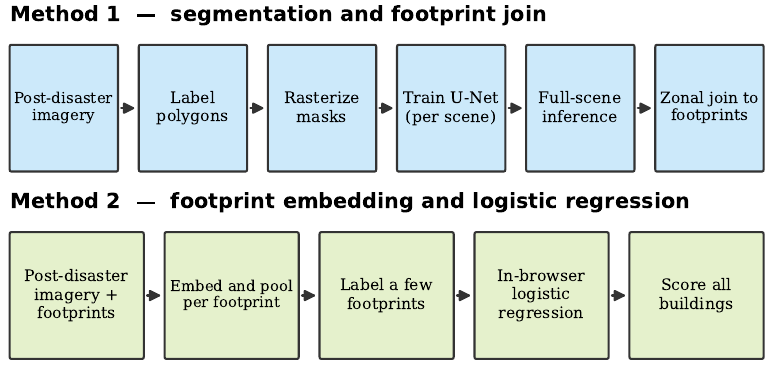}
\caption{The two HASTE methods.
Method 1 (top) labels polygons on the post-disaster image, trains a per-scene segmentation model, runs it over the whole scene, and joins the pixels to building footprints.
Method 2 (bottom) embeds and pools features over each footprint, has the analyst label a few buildings, and fits an in-browser logistic regression that scores every building.}
\label{fig:workflows}
\end{figure}

The first method treats damage as a segmentation problem on the post-disaster image and then summarizes the pixel predictions per building.
Figure~\ref{fig:workflows} (top) shows the flow.

\paragraph{Labeling and masks.}
The analyst draws polygons on the post-disaster scene for a small number of classes, typically background, intact building, and damaged building.
HASTE rasterizes these polygons into a per-pixel label mask and marks unlabeled pixels so that training ignores them.
To teach the model the boundary between buildings and their surroundings, it buffers the building class outward by a few meters and fills the buffer with the background class.
Because most pixels in a labeled scene remain unlabeled, HASTE follows the earlier workflow \citep{robinson2022fast} and clusters the sparse labels into a grid of small tiles, keeping only tiles that contain enough labeled pixels, so that training batches have labels rather than unlabeled background.

\paragraph{Per-scene training.}
HASTE trains a fresh model for each scene rather than reusing weights across events.
The model is a U-Net with a ResNeXt-50 encoder initialized from ImageNet \citep{ronneberger2015u,xie2017aggregated}, trained with cross-entropy on the labeled pixels and light augmentation, using the TorchGeo trainers \citep{stewart2025torchgeo}.
The goal is a model that fits this one image well, which sidesteps the domain shift that defeats a globally trained model.
Training does not drive the loss to zero, and the epoch budget is modest.
In operational use a full label, train, and review iteration completes in roughly one to two hours, so an analyst can run several rounds within a response window.

An optional constraint class helps when precise damage labels are expensive but an analyst can sweep and annotate large areas that are clearly undamaged.
The analyst marks such an area with a coarse polygon, and HASTE adds a term that penalizes any damaged prediction inside it.
Let $\mathcal{S}$ be the precisely labeled pixels and $\mathcal{C}$ the coarse ``known undamaged'' pixels.
The training loss is
\begin{equation}
\mathcal{L} = \frac{1}{|\mathcal{S}|}\sum_{i \in \mathcal{S}} \mathrm{CE}(\hat{y}_i, y_i)
            + \frac{1}{|\mathcal{C}|}\sum_{i \in \mathcal{C}} p_i(\text{damaged}),
\end{equation}
where $p_i(\text{damaged})$ is the softmax probability the model puts on the damaged class at pixel $i$.
The coarse pixels carry no per-pixel target and enter only through the penalty, which pushes the predicted damage probability toward zero across the marked area.
This turns cheap negative supervision into a regularizer against false positives.

\paragraph{Inference and footprint join.}
Once trained, the model runs over the whole scene in overlapping tiles, and HASTE takes the most likely class at each pixel to form a full-scene prediction.
It then summarizes these pixels per building.
For each footprint it reads the prediction pixels that fall inside the polygon and computes the fraction the model predicts as damaged,
\begin{equation}
d_b = \frac{\#\{\text{pixels in } b \text{ predicted damaged}\}}{\#\{\text{valid pixels in } b\}},
\end{equation}
at three buffer radii of 0, 10, and 20 meters around the footprint.
The buffers make the summary robust to small misalignment between the imagery and the footprints and to rubble that spills just outside the outline.
The result is a damage fraction per building rather than a hard label, which the report can threshold as needed.

\section{Method 2: footprint embedding and logistic regression}
\label{sec:method2}

The second method skips model training on the GPU.
It represents each building with a pooled feature vector from a pretrained model and fits a small classifier in the browser from a few labeled buildings.
Figure~\ref{fig:workflows} (bottom) shows the flow.

\paragraph{Embedding and pooling.}
HASTE runs a pretrained backbone over the post-disaster imagery to get a grid of patch-token features, then pools the tokens that fall inside each building footprint into one vector for that building.
The default backbone is MOSAIKS, which uses random convolutional features and is fast and cheap \citep{rolf2021generalizable}, and the platform also supports DINOv2 backbones \citep{oquab2023dinov2}.
Pooling is a masked mean over the tokens covered by the footprint, so a larger building averages over more tokens and each building ends with a fixed-length vector regardless of its size.
The experiments in Section~\ref{sec:experiments} use a richer pooling that concatenates per-channel statistics of the same tokens.
Buildings that fall outside the imagery keep a missing vector rather than losing their row, so every footprint has exactly one row and the join back to the map stays aligned.
HASTE writes the embeddings out in three forms: a table for analysis, vector tiles for streaming to the map, and a compact binary sidecar that the browser reads by row.

The method needs only the post-disaster image.
It does not depend on a clean pre-event image, which is its main practical advantage over the two-branch models.

\paragraph{In-browser classifier.}
As the analyst clicks buildings and labels them, HASTE fits a logistic regression on the labeled feature vectors, entirely in the browser.
It standardizes each feature, then minimizes the regularized logistic loss
\begin{equation}
\min_{w,b}\ \frac{1}{n}\sum_{i=1}^{n}
  \log\!\big(1 + e^{-y_i (w^\top x_i + b)}\big) + \frac{\lambda}{2}\lVert w \rVert^2 ,
\end{equation}
with $\lambda = 0.01$, by gradient descent for a few hundred steps.
Training starts once the analyst has labeled at least three buildings across at least two classes, and the model re-fits and re-scores the visible buildings as more labels come in, so the analyst sees the effect of each label immediately.
The model handles more than two damage states one-versus-rest.
The same math runs as a WebGPU shader when the browser supports it, with a plain CPU fallback, which keeps scoring interactive even for scenes with hundreds of thousands of buildings.
At 1024 features per building the shader scores roughly 100{,}000 buildings per second on an integrated laptop GPU and over a million on an Apple-silicon Mac, fast enough to re-score the visible viewport as the analyst labels; \ref{app:webgpu} reports throughput on two systems.

\paragraph{Scoring the scene.}
When the analyst is satisfied, HASTE scores every building in the scene and writes the predictions back onto the footprints as the same damage field that Method 1 produces.
From there the shared validation and assessment reports work without change, so the two methods are interchangeable from the report's point of view.

The two methods trade labeling effort against compute and turnaround.
Method 1 needs polygon labels and a GPU training job, and it produces a pixel-level map that shows damage that falls between or outside footprints, but a single relabel-and-review pass takes an hour or two.
Method 2 needs only a few clicked buildings and no training job, and it retrains and re-scores in seconds, so the analyst can iterate many times in the time Method 1 takes for one pass, at the cost of labeling and predicting only at the level of known footprints.

\section{Preliminary experiments}
\label{sec:experiments}

To understand how the embedding method behaves and which choices matter, we ran a set of experiments on xBD \citep{gupta2019xbd} in the post-disaster-only footprint setting that Method 2 uses.
We concentrate on Method 2 because its in-browser loop is where iteration happens during a response (Section~\ref{sec:method2}), and because its design choices, which embedding, how to pool, and how many labels it needs, are the ones these experiments can directly inform.
The experiments are preliminary, meant to guide the design of the platform rather than to serve as a published benchmark, and we report them for Method 2 alone; a matched evaluation of Method 1 is future work.

\paragraph{Setup.}
We collapse the xBD damage scale to a binary target, treating minor, major, and destroyed as damaged and no-damage as intact, and we drop unclassified buildings.
For each embedding backbone we pool the per-channel minimum, maximum, mean, and standard deviation of the tokens over every footprint and fit a classifier per disaster.
We split by scene so that the buildings used for testing come from scenes the classifier never saw during training, and we report ROC-AUC averaged with equal weight across disasters.
Where a result reports a spread, it is the mean and standard deviation over five random draws of the training labels, each draw standing in for a different set of buildings an analyst might label.

\paragraph{RQ1: How does accuracy scale with the amount of labeled data?}
Table~\ref{tab:fewshot} sweeps six backbones over per-disaster label budgets from 1\% to 50\% of the training buildings, with five random label draws each.
Accuracy climbs and then flattens: the strongest backbones reach 0.84 macro ROC-AUC from 1\% of the labels, a few dozen buildings per disaster, and about 0.92 by 50\%.
MOSAIKS, built from random rather than pretrained convolutional features, trails every self-supervised backbone across the sweep by a wide margin, which is why HASTE offers the pretrained backbones alongside it.
A second observation also shaped the platform: backbone size matters less than pretraining, since the small DINOv3 variant stays within about a point of the large one across the sweep.

\input{tables/ablations_fewshot.tex}

\paragraph{RQ2: How does it compare with supervised fine-tuning?}
As a reference for the standard way of approaching xBD, we fine-tune a single ResNet-50 end to end on post-event crops of the training buildings, shared across disasters (last row of Table~\ref{tab:fewshot}).
The embeddings are far more label-efficient.
With 1\% of the labels the strongest backbones reach 0.84 macro ROC-AUC while the fine-tuned network trained on the same 1\% reaches 0.77; at 5\% the embeddings match the fine-tuned network trained on every label, and at 10\% they pass it, 0.91 against 0.88.

\section{Operational responses}
\label{sec:responses}

HASTE and the workflows that preceded it have supported 31 distinct events since early 2023, spanning earthquakes, hurricanes, cyclones, floods, wildfires, and tornadoes.
Results reached partners including the American Red Cross, World Central Kitchen, the World Food Programme, the United Nations Development Programme (UNDP), and the United Nations Office for the Coordination of Humanitarian Affairs (OCHA) within hours to days of imagery becoming available.
The Microsoft AI for Good Research Lab maintains a public catalog of most of these responses as before-and-after visualizers with the resulting damage layers, currently 45 visualizers built from commercial optical imagery and from Sentinel-1 radar where cloud blocked the optical view.
It releases the damage layers openly through the \href{https://data.humdata.org/organization/microsoft-ai4g-lab}{Humanitarian Data Exchange}.

A few responses show the range.
After the February 2023 Turkey earthquakes, working with the national disaster agency, the lab mapped four cities in the southeast from imagery in the first three days, estimating 3,800 damaged buildings and, by overlaying WorldPop population data \citep{tatem2017worldpop}, 160,000 people affected \citep{robinson2023turkey}.
The March 2023 Rolling Fork tornado response, with the American Red Cross, trained a per-event model that reached 0.86 precision and 0.80 recall against field ground truth in under two hours per scene \citep{robinson2023rapid}.
For the August 2023 Maui wildfire, SkySat imagery of Lahaina became available for download at 9\,am Pacific time and the lab delivered the assessment by 1\,pm, identifying 1,700 damaged buildings, at least 1,200 of them destroyed, and validating at roughly 97\% accuracy with 99\% recall and 96\% precision.
In January 2025 the lab mapped the Palisades and Eaton fires in Los Angeles, and it has since responded to the Myanmar earthquake in March 2025, tropical cyclones in Sri Lanka and Sumatra, and earthquakes along the Venezuelan coast in 2026.

\begin{figure}[t!h]
\centering
\includegraphics[width=0.9\textwidth]{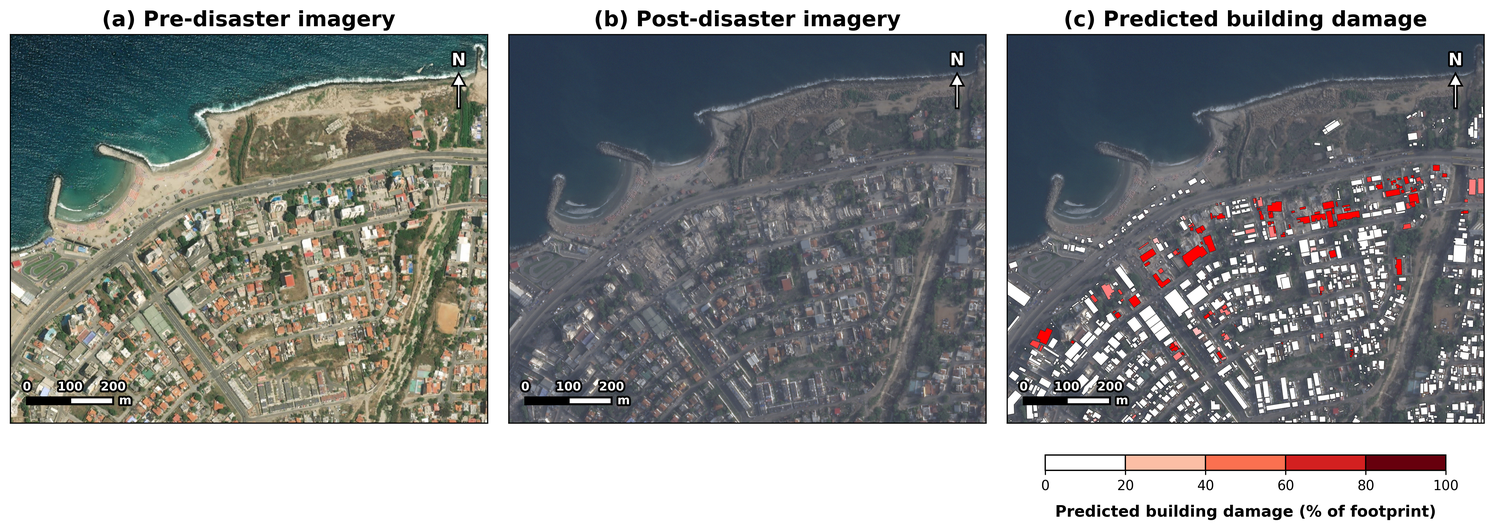}
\includegraphics[width=0.9\textwidth]{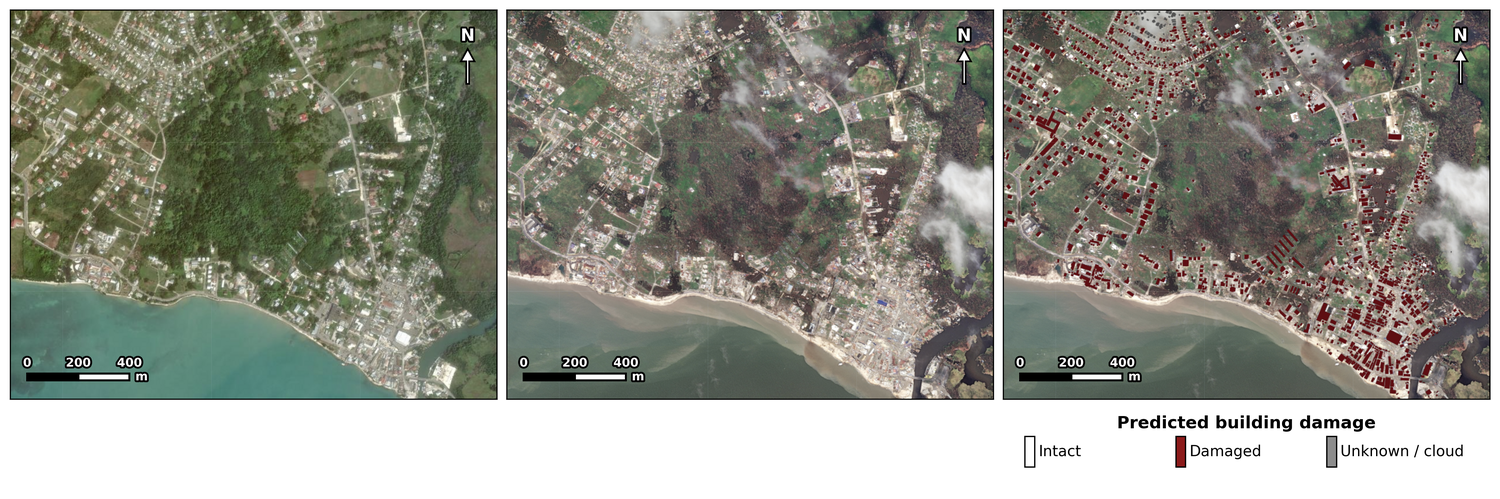}
\caption{HASTE rapid building-damage assessment across two disaster responses.
Each row shows, from left to right: (a) pre-disaster high-resolution satellite imagery, (b) post-disaster high-resolution satellite imagery, and (c) HASTE-predicted per-building damage.
Top row: the 2026 Venezuela earthquakes response showing Playa Los Corales area in Caraballeda, La Guaira, Venezuela, where damage is represented as the estimated damaged fraction of each building footprint, ranging from 0--20\% to 80--100\% from light to dark red.
Bottom row: Hurricane Melissa response in Jamaica, where each building is classified as intact, damaged, or unknown/cloud-obscured.
All panels are north-up; scale bars denote 200\,m for the top row and 400\,m for the bottom row.}
\label{fig:venezuela}
\end{figure}

Two recent responses show the validation reports of Section~\ref{sec:tool} in use.
After Hurricane Melissa made landfall in Jamaica on 28 October 2025, the lab assessed four areas of interest covering about 2,300 square kilometers, using post-disaster Vantor satellite imagery over Black River and Montego Bay and NOAA/NGS aerial imagery over the St.\ Elizabeth and Westmoreland parishes.
The workflow trained per-scene segmentation models (Method 1) to classify pixels as building, damaged, cloud, or background, joined the predictions to Overture Maps building footprints, and released per-footprint GeoPackage outputs with damage and cloud/unknown attributes.
Cloud cover was a major operational constraint in the satellite scenes, obscuring 65,000 of 110,000 footprints in Black River and 17,000 of 61,000 in Montego Bay.
Among the assessable footprints, HASTE identified widespread damage, including buildings that could no longer be detected in the imagery, and independent validation samples estimated 96\% recall and 82\% precision in Black River and 86\% recall and 71\% precision in Montego Bay.
Extrapolating validation performance to buildings larger than 50 square meters, the reports estimated 31,000 damaged buildings in Black River, 62\% of those large enough to assess (95\% confidence interval 26,000 to 37,000), and 18,000 in Montego Bay, 49\% (95\% confidence interval 15,000 to 20,000).
The aerial assessments further estimated 20,000 damaged buildings in St.\ Elizabeth and 21,000 in Westmoreland, with the St.\ Elizabeth validation sample showing 76\% recall and 91\% precision.
This response highlights why the reports expose validation metrics, cloud and unknown flags, and confidence intervals rather than presenting automated damage layers as ground truth.

Cyclone Gezani in Toamasina, Madagascar, shows the same workflow used for repeat assessment after an event.
The lab first assessed 90 square kilometers of Airbus imagery from 15 February 2026, then ran a follow-up assessment using 40 square kilometers of Vantor imagery from 29 March 2026 over parts of the same city.
Both passes used the Method 1 workflow.
The February assessment estimated 23,000 damaged buildings larger than 50 square meters, 55\%, with 97\% recall and 79\% precision; the March follow-up estimated 13,000 damaged buildings, 43\%, with 83\% recall and 87\% precision.
Joining the two dates by common cloud-free footprints showed that many buildings classified as damaged in February were classified as undamaged in March, suggesting repair, cleanup, changing visibility, or model disagreement between passes.
The case shows that repeated assessments can surface recovery, though separating true change from model disagreement requires co-registered passes and validated error rates.

Following the June 2026 earthquakes along the Venezuelan coast, the lab used HASTE to map building damage across the La Guaira coastal region, including Catia La Mar, La Guaira, and Caraballeda (Figure~\ref{fig:venezuela}).
Five post-disaster optical scenes from Planet, Vantor, and BlackSky covered roughly 210 square kilometers, and the same per-scene workflow classified pixels and joined the predictions to pre-event Overture footprints \citep{overture2024}.
Of about 72,000 buildings, 3,000 were obscured by cloud and marked unknown; 8,400 were flagged as damaged, 12\% of the cloud-free ones.
Where scenes overlapped, we deduplicated observations by spatial matching and kept disagreement between scenes as a per-building uncertainty attribute for downstream review.

\begin{figure}[t]
\centering
\includegraphics[height=2in]{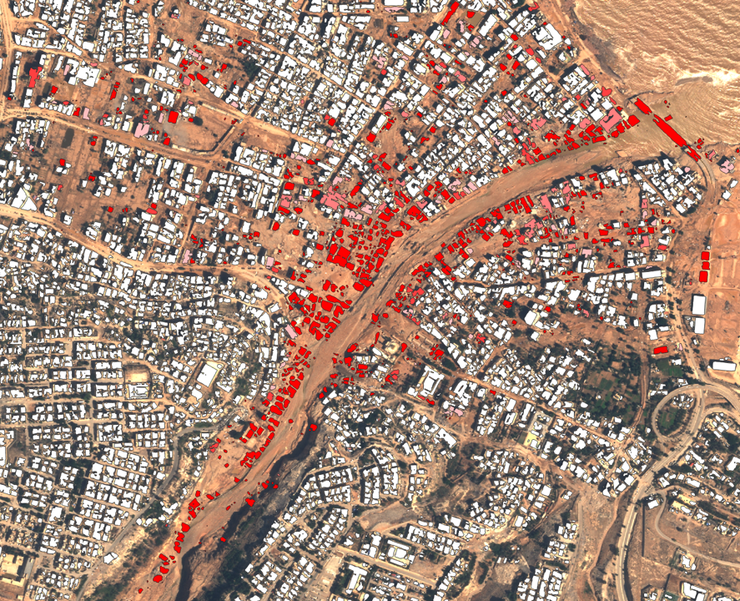}\hfill
\includegraphics[height=2in]{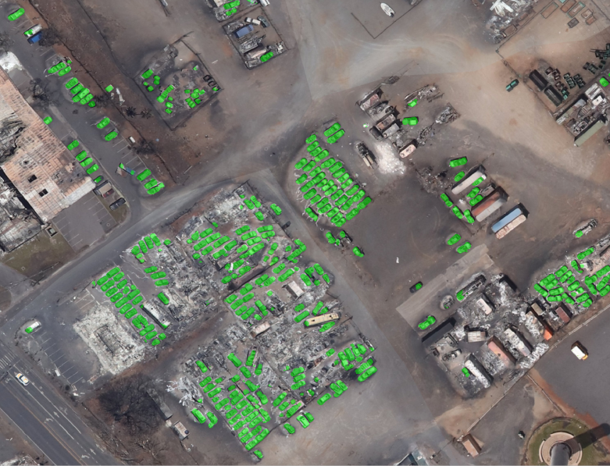}
\caption{The per-scene workflow adapts beyond building damage.
Left, flood extent and damaged infrastructure mapped in Derna, Libya after the September 2023 dam failures.
Right, burned vehicles detected in Lahaina, Maui to support debris-removal volume estimates after the August 2023 wildfire.}
\label{fig:beyond}
\end{figure}

The lab has also applied the same workflow to targets other than buildings (Figure~\ref{fig:beyond}).
After the September 2023 dam failures in Derna, Libya, it delineated flood extent and damaged infrastructure, and in Lahaina it detected burned vehicles in support of debris-removal volume estimates.
Each of these adaptations reused the labeling, training, and review cycle unchanged, only with different label classes.

External evaluation gives a sense of where these maps stand.
Ramachandran et al.\ overlaid the Turkey damage product on health-facility locations and validated against ground truth \citep{ramachandran2025implementation}.
They report high specificity, at 94\%, meaning the map rarely flags an undamaged facility, together with low sensitivity, at 43\%.
Agreement between the Turkey product and a comparable Google model was fair, with a Cohen's kappa of 0.32.
Aggregating predictions over an area raises sensitivity to 71\% and the kappa to 0.38, at some cost in specificity.
This matches a limitation the original report noted, that the footprint join does not credit rubble falling outside a building's outline to that building, which makes the map conservative and prone to miss some damaged structures.

\section{Limitations}
\label{sec:limitations}

HASTE's maps are prioritization aids for response and recovery, not a replacement for field validation.
They estimate damage visible from overhead, so damage that does not show from above, such as interior damage under an intact roof, goes uncounted; they carry both false positives and false negatives; and they mark buildings as unknown where cloud, missing footprints, or misregistration leave the imagery ambiguous.
The external evaluation in Section~\ref{sec:responses} gives the direction of the error: because the footprint join does not credit rubble that falls outside a building's outline, the maps run conservative, with high specificity but low sensitivity.

Both methods also inherit the limits of the footprint layer.
Method 2 labels and predicts only at the level of known footprints, so damage between or outside them is invisible to it, and footprints drawn from a years-old basemap can be missing or misaligned with the response imagery, in which case the pooling reads the wrong pixels.
The validation reports of Section~\ref{sec:tool} are labeled by visual interpretation of the imagery rather than by field survey, so except where field ground truth exists, as at Rolling Fork, the reported precision and recall measure agreement with an expert reading of the scene.

Finally, the experiments in Section~\ref{sec:experiments} are preliminary.
They collapse the xBD damage scale to a binary target, cover Method 2 alone, and run on a benchmark that is better registered than most response imagery, so the label-efficiency numbers should guide the platform's design rather than promise operational accuracy.
Likewise, comparing repeated assessments of the same city, as in Toamasina, stays suggestive until the passes are co-registered and the model's error rates are accounted for.

\section{Future directions}
\label{sec:future}

Several directions would make HASTE faster, more accurate, or applicable to more of what a disaster damages.
We group them by theme.

\paragraph{Vision-language assessment.}
Large vision-language models can describe an image and answer questions about it, and models tuned for overhead imagery are beginning to appear \citep{kuckreja2024geochat}.
A model of this kind could label buildings from a short prompt, give a first-pass damage read before any human labeling, or explain why it marked a building as damaged, which would help an analyst trust and correct the output.
Our early work in this direction, GeoVision Labeler, uses a vision-language model to describe each tile and a language model to map that description to user-defined classes, and reaches 93\% zero-shot accuracy on a buildings-versus-background task without any task-specific training \citep{hacheme2025geovision}.
How reliable such models are for fine-grained damage, and how to combine their judgments with the small classifiers HASTE already trains, are open questions.

\paragraph{Active learning.}
HASTE is interactive, so it is well placed to ask the analyst for the labels that will help most \citep{settles2009active}.
Rather than leaving the first clicks to chance, the platform could rank the unlabeled buildings and surface a useful set to label next, by model uncertainty, by diversity in embedding space, or by a hybrid that starts with diversity and switches to uncertainty once the classifier has seen enough labels.
In early offline experiments, spreading the first labels across the embedding space beat random selection at small label budgets, and building the suggestion into the labeler is a near-term step.
The same machinery helps with imagery problems such as haze, where the model could flag the buildings it is least sure about in the affected area and let the analyst resolve them.

\paragraph{Damage-assessment foundation models.}
The experiments in Section~\ref{sec:experiments} show that a general backbone already gives useful features, and that the best backbone varies with the label budget.
A model pretrained specifically for damage assessment, across many past events and sensors, could raise the floor for every new response, especially for imagery types that are underrepresented today.
A related and cheaper option is scene-specific self-supervision at response time.
One concrete version would warm-start Method 1's segmentation model by training its encoder to match patch-token features from a frozen DINOv3 teacher \citep{simeoni2025dinov3} on the scene's own imagery before the analyst draws any labels; the same idea applied to Method 2's embeddings would let the features adapt to the response sensor rather than staying frozen.
Either would suit the wide variety of sensors HASTE has to handle.

\paragraph{Roads and other infrastructure.}
Buildings are only part of what a disaster damages.
Blocked or washed-out roads, damaged bridges, and downed infrastructure shape how aid moves, and the same imagery captures them.
Individual responses have already stretched the segmentation loop to flood extent and burned vehicles (Section~\ref{sec:responses}), but each adaptation was ad hoc.
Supporting these targets as standard options in HASTE would widen the platform's use without changing its design, since the pooling and per-feature classification in Method 2 apply to any geometry.

\paragraph{Robustness and registration.}
Two smaller improvements would help both methods.
When the imagery and the footprints are misaligned, Method 2 pools over the wrong pixels, so learning a small alignment at the same time as the classifier could recover accuracy at little cost.
This matters most for response imagery joined to footprints from a years-old basemap, where misregistration is common, rather than for a well-registered benchmark like xBD.
Handling haze and cloud, either by masking affected pixels with a cloud detector \citep{wright2025training} or by giving the classifier features that let it model the haze, would make the post-only setting more reliable.

\section{Conclusion}

HASTE turns several years of one-off disaster-response pipelines into a no-code platform that partner organizations can run themselves.
It avoids the domain shift that limits globally trained damage models by training a small model for each event from a few local labels.
It offers two ways to do so, one based on per-scene segmentation and one on footprint embeddings with an in-browser classifier, and the two trade labeling effort for turnaround.
On xBD, the embedding route with a tenth of the labels outperforms a supervised network fine-tuned on all of them, which supports building the platform around frozen representations and small classifiers.
The platform and its predecessors have already supported more than thirty real responses across many disaster types and imagery sources.
The methods are simple by design, and the next steps follow from that simplicity: better representations, interactive labeling that asks for the most useful labels, and coverage of the roads and infrastructure that damage assessment has so far left out.

\section*{Acknowledgments}

We thank our partners in the humanitarian and disaster-response community, including national disaster management agencies, the American Red Cross, and the imagery providers whose data made these responses possible.

\printbibliography

\clearpage
\appendix

\section{In-browser scoring throughput}
\label{app:webgpu}

Table~\ref{tab:webgpu} reports the throughput of Method 2's in-browser classifier, the number of buildings scored per 0.1\,s, as the feature-vector length grows, for the WebGPU shader and the CPU fallback on two systems.
Two patterns hold on both: throughput falls as the feature vector grows, and the shader's advantage over the CPU widens with it.
The size of that advantage depends on the GPU.
On a Windows laptop with an integrated Adreno X1-85 the shader runs from parity at 256 features to 3.7$\times$ at 4096, while on an Apple M1 Max it runs from 7.9$\times$ at 256 features to 58.7$\times$ at 4096.
At 1024 features the shader scores about 100{,}000 buildings per second on the slower of the two systems and about 1.5 million on the faster, so even a scene of several hundred thousand buildings scores in seconds on either.

\begin{table}[h]\centering
\caption{In-browser scoring throughput on two systems: buildings scored per 0.1\,s (mean $\pm$ standard deviation over the listed runs) as a function of the feature-vector length, for the WebGPU shader and the CPU fallback.
Each run scores 2{,}048 buildings with a three-class one-versus-rest model under a 120\,ms budget; browsers do not expose exact CPU and GPU details.}
\label{tab:webgpu}
\small
\begin{tabular}{rcccc}
\toprule
Features/building & WebGPU (bldgs/0.1\,s) & CPU (bldgs/0.1\,s) & Speedup & Runs \\
\midrule
\multicolumn{5}{l}{\textit{Windows laptop: integrated Adreno X1-85, 12 logical cores, Chromium 150}} \\
256  & $18{,}058 \pm 656$     & $17{,}216 \pm 3{,}318$ & 1.0$\times$  & 5 \\
512  & $14{,}761 \pm 946$     & $8{,}792 \pm 1{,}050$  & 1.7$\times$  & 5 \\
1024 & $10{,}689 \pm 2{,}122$ & $4{,}324 \pm 233$      & 2.5$\times$  & 5 \\
2048 & $7{,}737 \pm 2{,}297$  & $2{,}310 \pm 65$       & 3.3$\times$  & 5 \\
4096 & $3{,}886 \pm 1{,}327$  & $1{,}062 \pm 299$      & 3.7$\times$  & 5 \\
\midrule
\multicolumn{5}{l}{\textit{Mac: Apple M1 Max, 10 logical cores, Chromium 150, macOS 15.7}} \\
256  & $188{,}635 \pm 28{,}740$ & $23{,}937 \pm 797$   & 7.9$\times$  & 5 \\
512  & $174{,}370 \pm 6{,}179$  & $11{,}367 \pm 31$    & 15.3$\times$ & 5 \\
1024 & $145{,}607 \pm 5{,}815$  & $5{,}274 \pm 117$    & 27.6$\times$ & 5 \\
2048 & $112{,}365 \pm 3{,}418$  & $2{,}606 \pm 31$     & 43.1$\times$ & 5 \\
4096 & $75{,}133 \pm 2{,}065$   & $1{,}280 \pm 32$     & 58.7$\times$ & 5 \\
\bottomrule
\end{tabular}
\end{table}

\end{document}

%% file: tables/ablations_fewshot.tex
\begin{table}[t]\centering
\caption{Macro ROC-AUC on xBD as the per-disaster label budget grows, mean $\pm$
standard deviation over five random label draws. All embedding rows use post-event
imagery only, statistics pooling, and a logistic-regression head tuned by
cross-validation; evaluation covers every labeled building in the held-out test
scenes. The last row is a single ResNet-50 shared across disasters and fine-tuned
end to end on post-event crops of each building. Bold marks the best value in each
column, ties included.}
\label{tab:fewshot}
\scriptsize
\setlength{\tabcolsep}{3pt}
\begin{tabular}{l ccccccc}
\toprule
Model & 1\% & 2\% & 5\% & 10\% & 25\% & 50\% & 100\% \\
\midrule
MOSAIKS               & $0.70_{\pm.012}$ & $0.73_{\pm.008}$ & $0.76_{\pm.011}$ & $0.80_{\pm.012}$ & $0.83_{\pm.005}$ & $0.84_{\pm.006}$ & -- \\
DINOv2 ViT-S/14       & $0.82_{\pm.014}$ & $0.85_{\pm.019}$ & $0.87_{\pm.015}$ & $0.89_{\pm.013}$ & $0.91_{\pm.006}$ & $\mathbf{0.92}_{\pm.003}$ & -- \\
DINOv2 ViT-B/14       & $0.82_{\pm.013}$ & $0.85_{\pm.014}$ & $0.87_{\pm.020}$ & $0.88_{\pm.017}$ & $0.90_{\pm.008}$ & $0.91_{\pm.004}$ & -- \\
DINOv3 ViT-S/16       & $0.83_{\pm.018}$ & $0.86_{\pm.020}$ & $0.88_{\pm.018}$ & $0.90_{\pm.006}$ & $0.91_{\pm.003}$ & $0.91_{\pm.003}$ & -- \\
DINOv3 ViT-L/16       & $\mathbf{0.84}_{\pm.021}$ & $\mathbf{0.87}_{\pm.022}$ & $0.88_{\pm.019}$ & $0.89_{\pm.007}$ & $0.91_{\pm.008}$ & $\mathbf{0.92}_{\pm.004}$ & -- \\
DINOv3-SAT ViT-L/16   & $\mathbf{0.84}_{\pm.020}$ & $\mathbf{0.87}_{\pm.024}$ & $\mathbf{0.89}_{\pm.016}$ & $\mathbf{0.91}_{\pm.005}$ & $\mathbf{0.92}_{\pm.004}$ & $\mathbf{0.92}_{\pm.004}$ & -- \\
\midrule
ResNet-50 fine-tune   & $0.77_{\pm.011}$ & -- & -- & -- & -- & -- & $0.88_{\pm.006}$ \\
\bottomrule
\end{tabular}
\end{table}